\def\year{2022}\relax
\documentclass[letterpaper]{article} 
\usepackage{aaai22}  
\usepackage{times}  
\usepackage{helvet}  
\usepackage{courier}  
\usepackage[hyphens]{url}  
\usepackage{graphicx} 
\usepackage{natbib}  
\usepackage{caption} 
\DeclareCaptionStyle{ruled}{labelfont=normalfont,labelsep=colon,strut=off} 
\usepackage{algorithm}
\usepackage{algorithmic}

\usepackage{newfloat}
\usepackage{listings}
\usepackage{multirow}

\floatstyle{ruled}
\newfloat{listing}{tb}{lst}{}
\floatname{listing}{Listing}
\nocopyright
\title{Community-Driven Comprehensive Scientific Paper Summarization:\\Insight from cvpaper.challenge}
\author{
    Shintaro Yamamoto\textsuperscript{\rm 1,2}\equalcontrib,
    Hirokatsu Kataoka\textsuperscript{\rm 2}\equalcontrib,
    Ryota Suzuki\textsuperscript{\rm 2},\\
    Seitaro Shinagawa\textsuperscript{\rm 3},
    Shigeo Morishima\textsuperscript{\rm 4}
}
\affiliations{
    \textsuperscript{\rm 1}Waseda University\\
    \textsuperscript{\rm 2}National Institute of Advanced Industrial Science and Technology (AIST)\\
    \textsuperscript{\rm 3}Nara Institute of Science and Technology\\
    \textsuperscript{\rm 4}Waseda Research Institute for Science and Engineering\\

}

\usepackage{bibentry}

\begin{document}

\maketitle

\begin{abstract}
The present paper introduces a group activity involving writing summaries of conference proceedings by volunteer participants.
The rapid increase in scientific papers is a heavy burden for researchers, especially non-native speakers, who need to survey scientific literature.
To alleviate this problem, we organized a group of non-native English speakers to write summaries of papers presented at a computer vision conference to share the knowledge of the papers read by the group.
We summarized a total of 2,000 papers presented at the Conference on Computer Vision and Pattern Recognition, a top-tier conference on computer vision, in 2019 and 2020.
We quantitatively analyzed participants' selection regarding which papers they read among the many available papers.
The experimental results suggest that we can summarize a wide range of papers without asking participants to read papers unrelated to their interests.
\end{abstract}

\section{Introduction}
Reading scientific papers is essential for researchers to acquire ideas and methods.
In the field of computer vision, the number of scientific papers accepted by leading conferences has been rapidly increasing.
In 2021, 1,661 and 1,617 were accepted to CVPR and ICCV, respectively.
Consequently, researchers need to read more papers, which is difficult to accomplish.
This is true especially for non-native speakers, who also have a language barrier to overcome when reading papers. Thus, there is a great demand for material that enables one to conduct a broad survey of research trends quickly and efficiently.
To address this problem, it is crucial to construct a cooperative framework to effectively grasp cutting-edge technologies from many papers.

The community cvpaper.challenge is a cross-institutional group of researchers and students in the field of computer vision in Japan.
To grasp trends in a research topic and present future directions, cvpaper.challenge leads a group activity involving writing summaries of papers presented at top-tier conferences, including CVPR, ICCV, and ECCV. [Define?]
Around 100 volunteers write summaries of computer vision papers in Japanese, and we make them publicly available\footnote{\url{https://xpaperchallenge.org/cv/survey/}}.
We can obtain overviews of conference papers from summaries, while reading papers by an individual researcher alone is difficult.
Summaries are also helpful in that they are written in Japanese, which removes the language barrier for Japanese speakers.

Topic coverage is an important aspect for grasping research trends in a field.
We should investigate the research trends for various topics to understand the entire field.
However, the topic coverage of summaries is unclear in our summarization activity.
As participants in our activity join as volunteers, they prefer to read papers according to their own preferences and specialties.
To clarify topic coverage, we quantitatively analyzed paper selection using a document representation method.
In the present paper, we first describe the process of writing summaries and then analyze the participants' selection of papers.

We present the case of writing summaries of papers in the proceedings of CVPR, an annual conference on computer vision.
According to Google Scholar\footnote{\url{https://scholar.google.com/}}, the proceedings of CVPR is the fourth and top publication in all disciplines and Engineering \& Computer Science, respectively\footnote{As of 22 February, 2022}.
Accordingly, CVPR is one of the most important conferences in the computer vision field and has been attracting numerous researchers.
Several studies in natural language processing also collected summaries of scientific papers.
These corpora are intended for scientific paper summarization.
In contrast, we aim to investigate scientific papers under the condition that participants in our activity can read papers as they like.
We analyzed the selection of papers to understand the behavior of participants who joined as volunteers.
We obtained a representation of scientific papers using a natural language processing-based approach and investigated personal preferences.
In addition, we also clarified whether summaries can cover papers on a wide variety of topics.

\section{Related Work}
Several researchers have built corpora of summaries of scientific papers.
Human-generated summaries in computational linguistics were provided for CL-SciSumm shared tasks~\citep{Jaidka2016,Jaidka2019b}, but the corpus size was limited to 60 papers.
The SciSummNet corpus, which consists of 1,000 computational linguistics papers, was proposed by~\citet{Yasunaga2019}.
Five Ph.D.\ students or people with equivalent expertise in natural language processing read the abstract from the paper and citation sentences to write a summary. 
Those summaries were designed to build corpora for automatic summarization of scientific papers.

Instead of those designed summaries, several studies used existing resources to build a corpus.
\citet{Collins2017} used highlighted statements provided by authors to build a corpus of 10k computer science papers.
\citet{Cohan2018b} proposed two datasets of arXiv and PubMed papers.
They collected more than 100k papers and considered the abstracts of papers as ground-truth summaries, which are easy to obtain.
SCITLDR, a corpus of single-sentence summaries of computer science papers, was proposed by~\citet{Cahola2020}.
They collected summaries either by extracting a summary submitted by authors on OpenReview or rewriting peer review comments by undergraduate students in computer science.
TALKSUMM~\citep{Lev2019} was built by using the transcripts of presentations in natural language processing and machine learning venues.

The present study focused on manual summary writing, but it was not well-designed to build a dataset, such as an existing corpus, because it depends on volunteers' cooperation.
Our requests to them included only minimal constraints to encourage the participation of many volunteers.
Participants could perform the task as they liked, such as selecting which papers to summarize.
This enabled the collection of a large volume of voluntary summaries.
However, similar to the studies using existing resources, voluntary summaries are not well-designed to construct a corpus of scientific paper summarization.
We then needed to investigate the feasibility or potential problems of corpus construction.
This paper aims to analyze the participants' behavior to clarify whether summaries written by volunteers can cover various topics in the field.

\section{Summary Writing}
This section describes the process of writing summaries of CVPR papers by a group of volunteers.
The activity attempts to archive the summaries of papers accepted by CVPR.
We began the activity soon after the proceedings were available.

\textbf{Participants}
To collect as many summaries of CVPR papers as possible, we welcomed any individuals interested in computer vision.
We announced the project on social media (Twitter) and in a Slack workspace to Japanese speakers interested in computer vision.
The participants were not required to have prior experience or knowledge, but they were required to write the summaries in Japanese.
Consequently, more than 100 people joined this activity both in 2019 and 2020 (although not every participant submitted a summary).
The participants included university students (undergraduate and graduate), engineers, and researchers with Ph.D.\ degrees.
The number of participants who submitted at least one summary is summarized in Table~\ref{tab:number}.

\begin{table}[t]
\centering
\caption{Number of participants and summarized papers in the CVPR comprehensive survey. Individuals who submitted at least one summary were counted as participants.}
\begin{tabular}{l|ccc}
\hline
Parameter & 2019 & 2020 \\ \hline
No. of participants & 90 & 95 \\
No. of summarized & 951 & 1,049 \\
No. of accepted to CVPR & 1,294 & 1470 \\
\hline
\end{tabular}
\label{tab:number}
\end{table}

In 2019, we created groups of three or four participants based on their research experience, and randomly selected papers were allocated to each group.
The purpose of the random assignment was to ensure that the summarized papers covered a broad range of topics.
In 2020, we also considered participants' interests to organize them into groups. 
The topics were defined by referring to programs in CVPR and were re-organized by the activity organizers.
We used Slack as a communication tool in both years and prepared a general channel for announcements (e.g., schedules) and group channels for sharing paper information.

\textbf{Writing Procedure}
Each participant first selected a paper to read from a list of papers we prepared.
In 2019, participants chose only from papers assigned to their group, but they could consider all papers in 2020. Only one participant read each paper.
After reading the selected paper(s), participants filled in a form to provide a summary.
The form consisted of four major sections: Overview, Novelty, Experimental results, and Other comments (optional).
Each section was limited to 200 characters in Japanese\footnote{According to a translation agency in Japan, 400 Japanese characters correspond to 200 or 250 words in English. \url{https://www.excellet.co.jp/blog/2017/302/} Accessed 18 Oct. 2021.}.
Participants could provide representative figures and tables as optional but highly recommended material.
Participants could also provide a word or phrase tags for searching (e.g., object detection and face recognition).
Submitted summaries were shared among participants.
We provided the instructions for the writing process as a presentation slide.
The whole writing procedure took approximately 7 and 5 weeks in 2019 and 2020, respectively.

\textbf{Statistics}
Table~\ref{tab:statistics} shows the number of papers that individual participants read.
On average, each participant read approximately 10 papers within the period.
Whereas some participants read more than 50 papers, some participants read only 1 paper.
The maximum number of papers read by individual participants increased from 2019 to 2020.
In 2019, we divided the participants into groups and assigned papers to each group, which restricted the paper selection.
In contrast, we did not assign papers in 2020 because several participants in 2019 mentioned that they could not read the papers they preferred.
As a consequence of the relaxed constraints, some participants might have read more papers without losing motivation.

\begin{table}[t]
\centering
\caption{Statistics for the number of papers read by individual participants.}
\begin{tabular}{l|cccc}
\hline
Year &Max  &Min  &Mean  &Median  \\ \hline
2019 &51  &1  &10.6$\pm$8.4  &9  \\
2020 &65  &1  &11.0$\pm$12.6  &7  \\ \hline
\end{tabular}
\label{tab:statistics}
\end{table}

\section{Analysis}
CVPR accepted more than 1,000 papers each year in 2019 and 2020.
Because an individual could read only a limited number of papers, covering  a wide variety topics by an individual participant was difficult.
Indeed, individual participants read at most 65 papers each year (Table~\ref{tab:statistics}).
We then investigated whether the resulting summaries covered a wide variety of topics in the field.

\textbf{Methodology}
We analyzed the paper selection both on individual and overall levels.
At the individual level, we investigated the paper selection by individual participants.
We analyzed the papers read by the five participants who read the most papers each year (Table~\ref{tab:top5}) to ensure that individual participants read a sufficient number of papers to analyze.
Note that some participants were ranked in the top five in both years, whereas other readers were analyzed for either in 2019 or 2020.
The top five participants wrote 17.7\% and 25.6\% of the summaries in 2019 and 2020, respectively.
At the overall level, we analyzed the paper selection by all participants.
We verified whether each paper was read by one of the participants or remained unread.

We analyzed the paper selection using the representation of a paper.
We used SPECTER~\citep{Cohan2020}, which is a state-of-the-art method to obtain a paper representation from a title and abstract.
SPECTER was trained so that the representation of a paper was similar to one that cites the paper.
We extracted feature vectors from all of the papers accepted by CVPR in 2019 and 2020. 

\begin{figure}
    \centering
    \includegraphics[width=\linewidth]{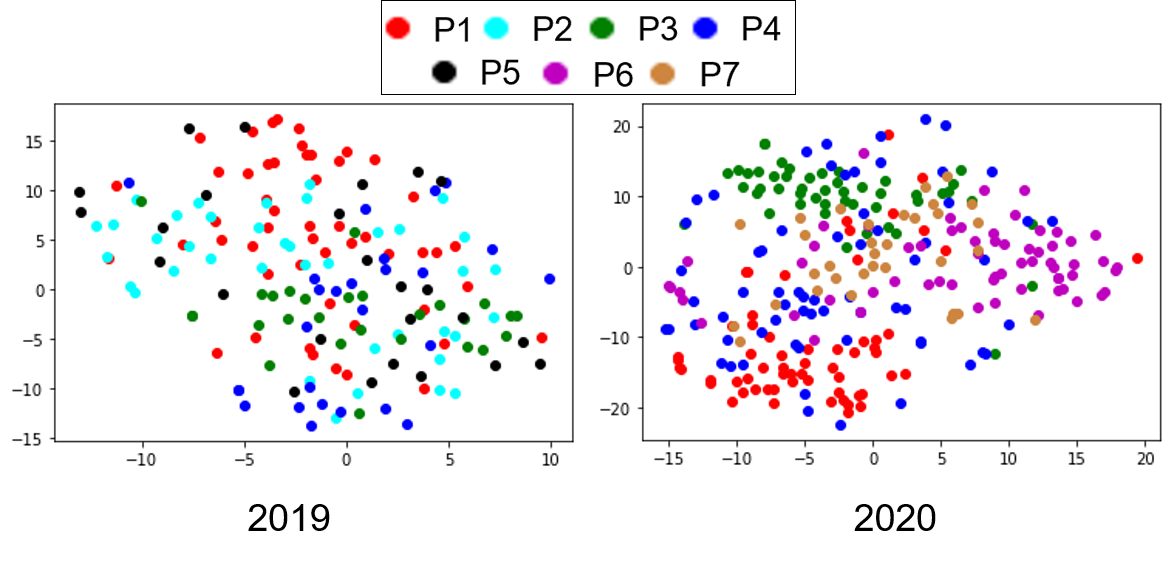}
    \caption{t-SNE visualization of SPECTER embeddings for each participant.}
    \label{fig:specter}
\end{figure}

\textbf{Individual Level}
A paper representation from SPECTER~\citep{Cohan2020} is a 768-dimensional vector, and we visualized feature vectors using t-distributed stochastic neighbor embedding (t-SNE) (Figure~\ref{fig:specter}).
In 2019, paper representations did not show individual tendencies.
We assigned participants to groups without considering personal preference in research topics in 2019.
In addition, each group was allocated papers randomly, resulting in participants being assigned papers outside their topics of interest.
To respond to participant requests, we changed the procedure so that group assignment was conducted based on self-reported preference in research topics, and participants could select any paper.
As expected, paper representations for some participants showed more personal tendencies in 2020, as compared with 2019.

\begin{table}[t]
    \centering
    \caption{Five participants who read the greatest number of papers in 2019 and 2020. The same participant IDs are used for participants who ranked in the top five in both years.}
    \begin{tabular}{l|ccc}
    \hline
    Year            &Rank  & Participant ID & No. of Papers read \\ \hline
    \multirow{5}{*}{2019}&1 & P1 & 51   \\
                      &2& P2 & 39  \\
                      &3& P3 & 27  \\
                      &4& P4 & 26  \\
                      &5& P5 & 25   \\ \hline
    \multirow{5}{*}{2020}&1 & P4 & 65   \\
                      &2& P6 & 63  \\
                      &3& P1 & 60  \\
                      &4& P3 & 48  \\
                      &5& P7 & 33  \\ \hline
    \end{tabular}
    \label{tab:top5}
\end{table}

To quantitatively verify whether participants tended to read papers of interest, we computed the distance between the distributions of papers that each participant read.
Here, we computed the earth mover’s distance (EMD)~\citep{Rubner2000}, which measures the similarity between distributions of paper representations.
To compare the difference between 2019 and 2020, we examined three participants who ranked in the top five in both years (P1, P3, and P4).
As shown in Table~\ref{tab:dist}, the EMDs for all pairs in 2020 are more significant than in 2019, which supports the idea that each participant read papers in their interest area in 2020.
Specifically, P1 and P3 show characteristic tendencies in paper selection in 2020 (Figure~\ref{fig:specter}), and therefore the EMD between these participants is much more prominent in 2020.
In contrast, P4 tended to read a wide variety of papers, even in 2020, which is consistent with the EMDs between P4 and the other participants being smaller.

\begin{table}[t]
    \centering
    \caption{Earth mover’s distance between distributions of papers read by individual participants.}
    \begin{tabular}{c|cc}
    \hline
    Participant pair & 2019 & 2020 \\ \hline
    P1, P3 & 59.7 & 69.7 \\
    P1, P4 & 61.5 & 64.3 \\
    P3, P4 & 62.5 & 63.1 \\ \hline
    \end{tabular}
    
    \label{tab:dist}
\end{table}

The personal tendency that paper representations of individual participants are concentrated in specific areas in 2020 indicates that participants would like to read papers on specific topics of interest.
Participant P1 used the tags related to vision and language, including ``vision and language’’ (13 times), ``image captioning’’ (5 times), ``visual question answering’’ (3 times), and ``visual dialog’’ (3 times).
Participant P3 also exhibited a personal preference.
A total of 19.4\% of tags used by P3 include the term ``point cloud,'' such as ``point cloud registration'' (5 times), ``point cloud segmentation'' (5 times), and ``point cloud classification'' (4 times).
Therefore, some participants selected papers from more than 1,000 available papers according to their interests.
However, the paper representations for P4 did not show personal tendencies, even for 2020 papers.
The tags used by P4 varied over a wide range of topics, for example, ``object detection'' (8 times), ``NAS'' or ``neural architecture search'' (6 times), and ``point cloud'' or ``3D point cloud'' (5 times), which refer to different topics of research in computer vision.
Thus, we can infer that some participants read papers covering various topics, while the others tended to select papers on specific topics.

\begin{figure}[t]
\centering
\includegraphics[width=\linewidth]{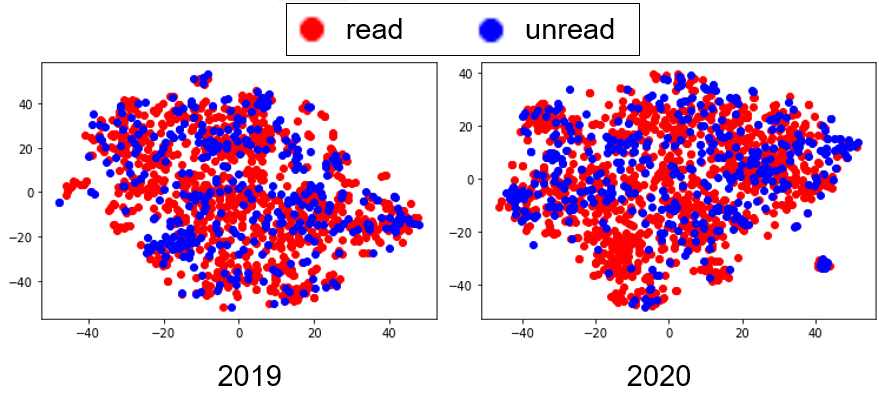}
\caption{t-SNE visualization of SPECTER embeddings for read and unread papers.}
\label{fig:unread}
\end{figure}

\begin{table*}[t]
\centering
\caption{Examples of ranking of papers read by one of the participants based on the Euclidean distance of SPECTER representation of unread papers (queries). Note that titles are from the proceedings of CVPR 2020.}
(a) Query: Neural Implicit Embedding for Point Cloud Analysis

\begin{tabular}{c|l}
\hline
Rank & \multicolumn{1}{c}{Title} \\ \hline
1     & PointGMM: A Neural GMM Network for Point Clouds      \\
2     & Learning to Segment 3D Point Clouds in 2D Image Space      \\
3     & Associate-3Ddet: Perceptual-to-Conceptual Association for 3D Point Cloud Object Detection      \\
\hline
\end{tabular}
\\
\vspace{5pt}
(b) Query: Learning Meta Face Recognition in Unseen Domains
\\
\begin{tabular}{c|l}
\hline
Rank & \multicolumn{1}{c}{Title} \\ \hline
1 & Towards Universal Representation Learning for Deep Face Recognition \\
2 & Domain Balancing: Face Recognition on Long-Tailed Domains \\
3&GroupFace: Learning Latent Groups and Constructing Group-Based Representations for Face Recognition\\
\hline
\end{tabular}

\label{tab:rank}
\end{table*}

\textbf{Overall Level}
It is essential to cover papers on various topics to survey papers in a research field.
Therefore, we clarified whether papers from a wide range of topics were summarized during the activity.
The distribution of feature vectors for read and unread papers is shown in Fig.~\ref{fig:unread}.
Paper representations indicate that papers read by one of the participants are widely distributed in the feature space.
This suggests that summaries generated by volunteers can cover a wide variety of topics in computer vision, even when individual participants read papers on a preferred topic.

We also verified whether the topic of a paper read by one of the participants is similar to that of an unread paper.
To validate this quantitatively, we ranked papers read by participants based on those papers' similarity to unread papers.
Given an unread paper as a query, papers read by one of the participants were ranked according to Euclidean distance.
Examples of the ranking results are shown in Table~\ref{tab:rank}.
Queries and papers ranked highly have shared terms in the title, such as ``point cloud'' and ``face recognition,'' indicating similar topics.
Papers on topics that are similar to those of unread papers were read by one of the participants, and therefore, a wide range of topics in computer vision is represented by summaries written with the help of groups of participants.

\section{Conclusion}
In this paper, we present a method for obtaining summaries of scientific publications by participants who read papers as part of an individual volunteer activity.
We obtained summaries of 2,000 papers that were presented at CVPR 2019 and 2020.
We then quantitatively analyzed the paper selection process.
In particular, freedom in paper selection is beneficial because we can cover a wide range of topics in the field, whereas participants could read papers of their interest without constraints.

\bibliography{aaai22}

\end{document}